\documentclass[10pt,twocolumn,letterpaper]{article}

\usepackage{cvpr}              

\usepackage{graphicx}
\usepackage{amsmath}
\usepackage{amssymb}
\usepackage{bm}
\usepackage{bbm}
\usepackage{booktabs}
\usepackage{xcolor}
\usepackage{pifont}
\newcommand{\cmark}{\ding{51}}%
\newcommand{\xmark}{\ding{55}}%

\usepackage[pagebackref,breaklinks,colorlinks]{hyperref}

\usepackage[capitalize]{cleveref}
\crefname{section}{Sec.}{Secs.}
\Crefname{section}{Section}{Sections}
\Crefname{table}{Table}{Tables}
\crefname{table}{Tab.}{Tabs.}

\def\cvprPaperID{7164} 
\def\confName{CVPR}
\def\confYear{2024}

\begin{document}

\title{Unsupervised Semantic Segmentation\\ Through Depth-Guided Feature Correlation and Sampling}

\author{Leon Sick\\
Ulm University\\
\and
Dominik Engel\\
Ulm University\\
\and
Pedro Hermosilla\\
TU Vienna\\
\and
Timo Ropinski\\
Ulm University\\
}
\maketitle

\begin{abstract}
   Traditionally, training neural networks to perform semantic segmentation requires expensive human-made annotations. But more recently, advances in the field of unsupervised learning have made significant progress on this issue and towards closing the gap to supervised algorithms. To achieve this, semantic knowledge is distilled by learning to correlate randomly sampled features from images across an entire dataset. In this work, we build upon these advances by incorporating information about the structure of the scene into the training process through the use of depth information. We achieve this by (1) learning depth-feature correlation by spatially correlating the feature maps with the depth maps to induce knowledge about the structure of the scene and (2) exploiting farthest-point sampling to more effectively select relevant features by utilizing 3D sampling techniques on depth information of the scene. Finally, we demonstrate the effectiveness of our technical contributions through extensive experimentation and present significant improvements in performance across multiple benchmark datasets.
\end{abstract}

\section{Introduction}
\label{sec:intro}
Semantic segmentation plays a critical role in many of today's vision systems in a multitude of domains. These include, among others, autonomous driving~\cite{feng2020deep}, medical applications~\cite{ronneberger2015u, wang2022medical}, and many more~\cite{liang2015semantic,xie2021segformer,zheng2021rethinking,cheng2021mask2former}. Until recently, the main body of research in this area was focused on supervised models that require a large amount pixel-level annotations for training. Not only is sourcing this image data often a labor intensive process, but also annotating the large datasets required for good performance comes at a high price. Several benchmark datasets report their annotation times. For example, the MS COCO dataset~\cite{lin2014microsoft} required more than 28K hours of human annotation for around 164K images, and annotating a single image in the Cityscapes dataset~\cite{cordts2016cityscapes} took 1.5 hours on average. These costs have triggered the advent of unsupervised semantic segmentation~\cite{ji2019invariant,cho2021picie,hamilton2021stego,seong2023hidden}, which aims to remove the need for labeled training data in order to train segmentation models.
Recently, work by Hamilton~\etal~\cite{hamilton2021stego} has accelerated the progress towards removing the need for labels to achieve good results on semantic segmentation tasks. Their model, STEGO, uses a DINO-pretrained~\cite{caron2021emerging} Vision Transformer (ViT)~\cite{dosovitskiy2020image} to extract features that are then distilled across the entire dataset to learn semantically relevant features, using a contrastive learning approach. The to-be-distilled features are sampled randomly from feature maps produced from the same image, k-NN matched images as well as other negative images. Seong~\etal~\cite{seong2023hidden} build on this process by trying to identify features that are most relevant to the model by discovering hidden positives. Their work exposes an inefficiency of random sampling in STEGO as hidden positives sampling leads to significant improvements. However, both approaches only operate on the pixel space and therefore fail to take into account the spatial layout of the scene. Not only do we humans perceive the world in 3D, but also previous work~\cite{hoyer2021three,wang2021domain,cardace2022plugging, vu2019dada, zhang2019pattern, chen2019towards} has shown that supervised semantic segmentation can benefit greatly from spatial information during training. Inspired by these observations, we propose to incorporate spatial information in the form of depth maps into the STEGO training process. Depth is considered a product of vision and does not provide a labeled training signal. To obtain depth information for the benchmark image datasets in our evaluations, we make use of
an off-the-shelf zero-shot monocular depth estimator to obtain spatial information of the scene. This allows us to incorporate the depth information without the need for human annotations or sensor ground truth.

\begin{figure}[!tb]
  \centering
   \includegraphics[width=\linewidth, page=1]{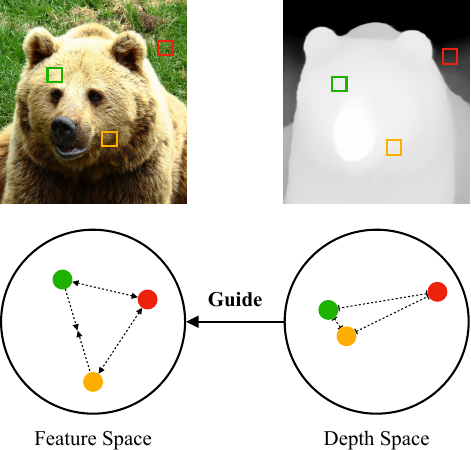}
   \caption{\textbf{Guiding the feature space for unsupervised segmentation with depth information.} Our intuition behind the proposed approach is simple: For locations in the 3D space with a low distance, we guide the model to map their features closer together. Vice versa, the features are learned to be drawn apart in feature space if their distance in the metric space is large.}
   \label{fig:intuition}
\end{figure}

With our method, \emph{DepthG}, we propose to \textbf{(1)} guide the model to learn a rough spatial layout of the scene, since we hypothesize this will aid the network in differentiating objects much better. We achieve this by extending the contrastive process to the spatial dimension: We do not limit the model to learning only Feature-Feature Correlations, but also \emph{Depth-Feature Correlations}. Through this process, the model is guided towards pulling apart the features with high distances in 3D space, as well as mapping them closer together if their distance is low depth space. Figure~\ref{fig:intuition} visualizes this process.

With the information about the spatial layout of the scene present, we furthermore propose to \textbf{(2)} spatially inform our feature sampling process by utilizing \emph{Farthest-Point Sampling (FPS)}~\cite{eldar1997farthest, qi2017pointnet++} on the depth map, which equally samples scenes in 3D. We show these techniques in combination make our approach to unsupervised segmentation highly effective, since in our evaluations across multiple benchmark datasets, we demonstrate state-of-the-art performance.
We include a short video presenting our method as part of the supplementary materials.
To the best of our knowledge, we are the first to propose a mechanism to incorporate 3D knowledge of the scene into unsupervised learning for 2D images \emph{without} encoding depth maps as part of the network input. This design aspect of our method alleviates the risk of the model developing an input dependency.
Therefore, our approach does not rely on depth information during inference and is not affected by availability or quality of such depth information.


%
%

\section{Related Work}

\subsection{Unsupervised Semantic Segmentation}
Recent works~\cite{cho2021picie, hamilton2021stego,ji2019invariant,seong2023hidden} have attempted to tackle semantic segmentation without the use of human annotations. Ji~\etal~\cite{ji2019invariant} propose IIC, a method that aims to maximize the mutual information between different augmented versions of an image. PiCIE, published by Cho~\etal~\cite{cho2021picie}, introduces an inductive bias based on the invariance to photometric transformations and equivariance to geometric manipulations. DINO~\cite{caron2021emerging} often serves as a critical component to unsupervised segmentation algorithms, since the self-supervised pre-trained ViT can produce semantically relevant features. Recent work by Seitzer~\etal~\cite{seitzer2023dinosaur} builds upon this ability by training a model with slot attention~\cite{locatello2020slotattention} to reconstruct the feature maps produced by DINO from the different slots. The features of their object-centric model are clustered with k-means~\cite{lloyd1982kmeans} where each slot is associated with a cluster. In their 2021 work, Hamilton~\etal~\cite{hamilton2021stego} have also built upon DINO features by introducing a feature distillation process with features from the same image, k-NN retrieved examples as well as random other images from the dataset. Their learned representations are finally clustered and refined with a CRF~\cite{krahenbuhl2011crf} for semantic segmentation. While STEGO's feature selection process is random, Seong~\etal~\cite{seong2023hidden} introduce a more effective sampling strategy by discovering hidden positives. During training, they form task-agnostic and task-specific feature pools. For an anchor feature, they then compute the maximum similarity to any of the pool features and sample locations in the image with greater similarity than the determined value. 
A more detailed introduction to STEGO is provided in Section \ref{sec:preliminary}.

\subsection{Depth For Semantic Segmentation}
Previous research~\cite{wang2021domain, hoyer2021three, cardace2022plugging, ying2022uctnet, hou2023mask3d, wang2022multimodal} has sought to incorporate depth for semantic segmentation in different settings. Wang~\etal~\cite{wang2021domain} propose to use depth for adapting segmentation models to new data domains. Their method adds depth estimation as an auxiliary task to strengthen the prediction of segmentation tasks. Furthermore, they approximate the pixel-wise adaption difficulty from source to target domain through the use of depth decoders. Work by Hoyer~\etal~\cite{hoyer2021three} explores three further strategies of how depth can be useful for segmentation. First, they propose using a shared backbone to share learning features for segmentation and self-supervised depth estimation, similar to Wang~\etal~\cite{wang2021domain}. Second, they use depth maps to introduce a data augmentation that is informed by the structure of the scene. And lastly, they detail the integration of depth into an active learning loop as part of a student-teacher setup. Another work by Hou~\etal~\cite{hou2023mask3d} incorporates depth information into a pre-training algorithm with the aim to learn better representations for semantic segmentation. In their work, Mask3D, they propose to incorporate depth in the form of a 3D prior by formulating a reconstruction task that operates on masked RGB and depth patches, enabling them to learn more useful features for semantic segmentation.


\section{Method}
\begin{figure*}
  \centering
  \includegraphics[width=\textwidth, page=2]{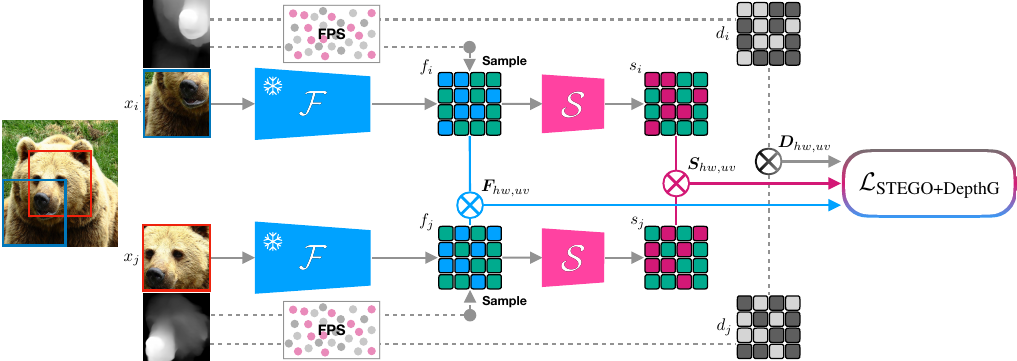}
  \caption{\textbf{Overview of the DepthG training process.} After 5-cropping the image, each crop is encoded by the DINO-pretrained ViT $\mathcal{F}$ to output a feature map. Using farthest point sampling (FPS), we sample the 3D space equally and convert the coordinates to select samples in the feature map. The sampled features are further transformed by the segmentation head $\mathcal{S}$. For both feature maps, the correlation tensor is computed. Following, we sample the depth map at the coordinates obtained by FPS and compute a correlation tensor in the same fashion. Finally, we compute our \emph{Depth-Feature Correlation} loss and combine it with the feature distillation loss from STEGO. We guide the model to learn depth-feature correlation for crops of the same image, while the feature distillation loss is also applied to k-NN-selected and random images.}
  \label{fig:main}
  \vspace{-3mm}
\end{figure*}

In the following, we detail our proposed method for guiding unsupervised segmentation with depth information. An overview of our approach is presented in Figure \ref{fig:main}.


\subsection{Preliminary}
\label{sec:preliminary}
Our approach builds upon work by Hamilton~\etal~\cite{hamilton2021stego}. In their work, each image is 5-cropped and k-NN correspondences between these images are calculated using the DINO ViT~\cite{caron2021emerging}. Generally, STEGO uses a feature extractor $\mathcal{F}: \mathbb{R}^{3\times H_\text{in}\times W_\text{in}} \rightarrow \mathbb{R}^{C\times H\times W}$ with input image height $H_\text{in}$ and width $W_\text{in}$, to calculate a feature map $f \in \mathbb{R}^{C\times H\times W}$ with height $H$, width $W$ and feature dimension $C$ from the input image. These features are then further encoded by a segmentation head $\mathcal{S} : \mathbb{R}^{C\times H\times W} \rightarrow \mathbb{R}^{Q\times H\times W}$ to calculate the code space $s \in \mathbb{R}^{Q\times  H\times W}$ with code dimension $Q$. With the goal of forming compact clusters and amplifying the correlation of the learned features, let $f_{i} := \mathcal{F}(x_i)$ and $f_{j} := \mathcal{F}(x_j)$ be feature maps for a given input pair of $x_i$ and $x_j$, which are then used to calculate $s_{i} := \mathcal{S}(f_{i})$ and $s_{j}:= \mathcal{S}(f_{j})$ from the segmentation head $\mathcal{S}$. In practice, STEGO samples $N^2$ vectors from the feature map during training. Hamilton~\etal~\cite{hamilton2021stego} introduce the concept of constructing the feature correspondence tensor $\bm{F} \in \mathbb{R}^{H\times W\times H\times W}$ as follows:
\begin{equation}
\bm{F}_{hw,uv} = \frac{f_i^{hw} \cdot f_j^{uv}}{\lVert f_i^{hw} \rVert \lVert f_j^{uv} \rVert}
\label{eq:feature_correlation}
\end{equation}
where $\cdot$ denotes the dot product. Using the same formula we obtain $\bm{S}$ using $s_i, s_j$. Consequently, the feature correlation loss is defined as:
\begin{equation}
\mathcal{L}_\text{Corr} := 
- \sum_{hw, uv} (\bm{F}_{hw,uv}- b) \max(\bm{S}_{hw,uv}, 0)        
\label{eq:feature_feature_correlation}
\end{equation}
where $b$ is a scalar bias hyperparameter.
Empirical evaluations from STEGO have shown that applying spatial centering to the feature correlation loss along with zero-clamping further improves performance~\cite{hamilton2021stego}.
    These correlations are calculated for two crops from the same image ($\mathcal{L}_\text{self}$)
 and one from a different but similar image, determined by the k-NN correspondence pre-processing ($\mathcal{L}_\text{knn}$). Finally, negative images are sampled randomly ($\mathcal{L}_\text{random}$).
The final loss is a weighted sum of the different losses where each of them has their individual weight $\lambda_i$:
\begin{equation}
  \mathcal{L}_\text{STEGO} = \lambda_\text{self} \mathcal{L}_\text{self} + \lambda_\text{knn} \mathcal{L}_\text{knn} + \lambda_\text{random} \mathcal{L}_\text{random}
  \label{eq:stego_loss}
\end{equation}
After training, the inferred feature maps for a test image are clustered using k-means and refined with a conditional random field (CRF)~\cite{krahenbuhl2011crf}.



\subsection{Depth Map Generation}

Since in many cases, depth information about the scene is not readily available, we make use of recent progress in the field of monocular depth estimation~\cite{bhat2023zoedepth,ranftl2020midas,bhat2021adabins,bhat2022localbins,lee2019big} to obtain depth maps from RGB images. Recently, methods from this field have made significant advances in zero-shot depth estimation, i.e. predicting depth values for scenes from data domains not seen during training. This property makes them especially suitable for our method, since it enables us to obtain high-quality depth predictions for a wide variety of data domains without ever re-training the depth network. This property also limits the computational cost for our method. 
For our method, we experiment with different state-of-the-art monocular depth estimators, detailed in Section~\ref{abl:depth-maps}, and found ZoeDepth~\cite{bhat2023zoedepth} to perform best in our evaluations. 
Given a cropped RGB image $x_i$, we use the monocular depth estimator $\mathcal{M}$ together with average pooling to predict depth $d_{i} \in [0,1]^{H\times W}$ at feature resolution:
\begin{equation}
  d_{i} = \text{pool}(\mathcal{M}(x_i))
  \label{eq:monodepth}
\end{equation}
The average pooling operation is used to match the dimensions of the feature map, which is required to sample non-overlapping locations at the patch resolution.


\subsection{Depth-Feature Correlation Loss}
With our \emph{Depth-Feature Correlation} loss, we aim to enforce spatial consistency in the feature map by transferring the distances from the depth map to the latent feature space.

In contrastive learning, the network is incentivized to decrease the distance in feature space for similar instances, therefore learning to map their latent representations closer together. Likewise, different instances are drawn further apart in feature distance. 
We assume the same concept to be true in 3D space: The spatial distance between two points from the same depth plateau is smaller, while the distance between a point in the foreground and one in the background is larger.
Since, in both spaces, the concept of measuring difference is represented by the distance between two points, we propose to align them through our concept of \emph{Depth-Feature Correlation}: For large distances in the 3D space, we encourage the network to produce vectors that are further apart, and vice versa. With this, we induce the model with knowledge about the spatial structure of the scene, enabling it to better differentiate between objects. To achieve this we construct the depth correspondence tensor similar to the feature correspondence from Equation~\ref{eq:feature_correlation}.
The depth correspondence tensor $\bm{D}$ is computed from the depths of two different image crops as follows:
\begin{equation}
    \bm{D}_{hw, uv} = d_{i}^{hw} d_{j}^{uv},
\label{eq:depth_correlation}
\end{equation}
where $(h, w)$ and $(u, v)$ represent the pixel positions in the depth maps $d_{i}$ and $d_{j}$ respectively.
Together with the zero clamping, our \emph{Depth-Feature Correlation} loss is defined as:
\begin{equation}
\mathcal{L}_\text{DepthG} = 
- \sum_{hw, uv} (\bm{D}_{hw, uv} - b_\text{DepthG}) \max(\bm{S}_{hw, uv}, 0)
\label{eq:depthg_loss}
\end{equation}
where \( \bm{D}_{hw, uv} \) represents the depth correlation tensor, \( b_\text{DepthG} \) is the bias for our loss term, and \( \bm{S}_{hw, uv} \) represents the feature correlation tensor computed from the output features of the segmentation head $\mathcal{S}$.
By also using zero-clamping, we limit erroneous learning signals that aim to draw apart instances of the same class if they have large spatial differences.
With this, we extend the STEGO loss so it can be formulated as follows:
\begin{equation}
  \mathcal{L}_\text{STEGO+DepthG} = \mathcal{L}_\text{STEGO} + \lambda_\text{DepthG} \mathcal{L}_\text{DepthG}
  \label{eq:full_loss}
\end{equation}
with \emph{Depth-Feature Correlation} weight $\lambda_\text{DepthG}$. By inducing depth knowledge during training \emph{without} encoding the depth maps as part of the model input, our model can predict spatially informed segmentations on RGB images with its distilled knowledge and does not rely on depth to be available at test time.


\subsection{Depth-Guided Feature Sampling}\label{sec:fps}
We also aim to make the feature sampling process informed by the spatial layout of the scene. To perform sampling in the 3D space, we transform the downsampled depth map $d(x_i)$ into a point cloud with points $\{p_1, p_2, ..., p_n\}$. On this point cloud, we apply farthest point sampling (FPS)~\cite{eldar1997farthest}, in an iterative fashion by always selecting the next point $p_{k}$ as the point with the maximum distance in 3D space with respect to the already sampled points $\{p_1, p_2, ..., p_{k-1}\}$. After having sampled $N^2$ points, we end up with a set of samples $\{p_1, p_2, ..., p_{N^2}\}$ which are consequently converted to 2D sampling indices for the feature maps $f$ and $g$. In contrast to the data-agnostic random sampling applied in STEGO, our feature selection process takes into account the geometry of the input scene and covers the spatial structure more equally.
In our ablations in Section~\ref{abl:influence}, we show this scene coverage from FPS of the depth space further increases the effectiveness of our \emph{Depth Feature Correlation} loss, due to the increased diversity in selected 3D locations. We show a visual comparison between random and farthest point sampling in Figure~\ref{fig:sampling}.


\subsection{Guidance Scheduling}
While our \emph{Depth-Feature Correlation} loss is effective at enriching the model's learning process with spatial information of the scene, we aim to alleviate the danger of it interfering with the learning of feature correlations during model training. We hypothesize that our model most greatly benefits from depth information in the beginning of training when its only knowledge is encoded in the features maps by the frozen ViT backbone. To give it a head start, we increase the weight of our \emph{Depth-Feature Correlation} loss in the beginning and gradually decrease its influence during training. Vice versa, the distillation process in the feature space will increasingly emphasised as the training progresses. In this way, the network builds upon the already learned rough spatial structure of the scene achieved through our depth guidance.
Therefore, we implement an exponential decay of the weight $\lambda_\text{DepthG}$ and bias $b_\text{DepthG}$ of our loss component. We ablate on the use of guidance scheduling in the appendix.


\subsection{Local Hidden Positives In 3D Space}\label{sec:lhp}
\begin{figure}[!tb]
  \centering
   \includegraphics[width=\linewidth, page=6]{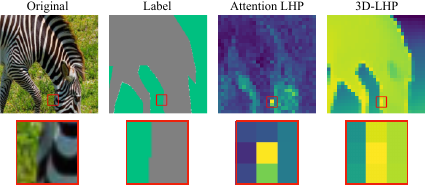}
   \caption{\textbf{Local Hidden Positives.} We visualize the use of depth and attention maps for local hidden positives. For this visualization, we sample the respective propagation maps at the yellow patch in the center of the crops. We observe the depth map to have sharper borders and more consistent propagation values. We experiment with both propagation strategies in Section~\ref{sec:lhp-abl}.}
   \label{fig:lhp}
\end{figure}

We further explore the combination of our method with Hidden Positives~\cite{seong2023hidden}, which also builds upon STEGO. As we demonstrate as part of our ablation on the individual influence of our contributions in Section~\ref{abl:influence}, our feature sampling method, implemented through farthest point sampling, is integral to the effectiveness of \emph{DepthG}. Therefore, we decide not to replace it with the global hidden positives sampling from Seong~\etal~\cite{seong2023hidden}. Instead, we implement a depth-informed variant of local hidden positives, \emph{3D-LHP}, where the loss for an individual patch is propagated to eight neighboring patches proportionally to their attention values obtained from the feature extractor. We modify this strategy by instead propagating the learning signal to the closest patches \emph{in 3D space} proportional to their relative distances, using the point cloud described in Section~\ref{sec:fps}. As visualized in Figure~\ref{fig:lhp}, we observe that our depth-based propagation map has sharper and more consistent surfaces. 
To propagate the learning signal to the selected patches, we follow Hidden Positives and mix the features coming from the segmentation head $\mathcal{S}$ in proportion to their propagation values (point distances). The mixed representations are then fed through an additional projection head $\mathcal{P}$. We then calculate $\mathcal{L}_\text{STEGO+DepthG}$ again for the produced output features and combine it with the loss from the segmentation head output.

\section{Experiments}
\vspace{-0.5mm}
\paragraph{Datasets and Models.} We conduct experiments on the COCO-Stuff~\cite{caesar2018coco}, Cityscapes~\cite{cordts2016cityscapes} and Potsdam-3 datasets. COCO-Stuff contains a wide variety of real-world scenes. In our evaluation, we follow~\cite{hamilton2021stego, seong2023hidden, ji2019invariant} and provide results on the coarse class split, COCO-Stuff 27. In contrast, Cityscapes contains traffic scenes from 50 cities from a driver-like viewpoint. Lastly, the Potsdam-3 dataset is composed of aerial, top-down images of the city of Potsdam. 
We use the DINO~\cite{caron2021emerging} backbones ViT-Small (ViT-S) and ViT-Base (ViT-B), which were pre-trained in a self-supervised manner on ImageNet-1k \cite{deng2009imagenet}. We choose the models with a patch size of $8\times 8$, since they have been shown to perform best for semantic segmentation due to the higher resolution of the resulting feature maps~\cite{hamilton2021stego, seong2023hidden}. In the result tables, we refer to \emph{DepthG} models trained with our \emph{Depth-Feature Correlation} loss and FPS as \textbf{Ours}. Models that utilize 3D-LHP propagation with depth maps in addition are displayed as \textbf{Ours w/ 3D-LHP}. We compare our approach against competing methods which were never trained with human annotations, i.e. human labels or language supervision, neither in the feature extractor nor the segmentation training.

\paragraph{Evaluation Protocols.} Similar to STEGO and related work~\cite{hamilton2021stego, seong2023hidden}, we evaluate our models in the unsupervised, clustering-based setting, as well as linear probing. Since the output of our model is a pixel-level map of features and not class labels, these features are consequently clustered. Following, the pseudo-labeled clusters are aligned with the ground truth labels through Hungarian matching~\cite{kuhn1955hungarian, kuhn1956variants} across the entire validation dataset. To perform linear probing, an additional linear layer is added on top of the model and trained with cross-entropy loss to learn classification of the features.



\begin{table}[!htb]
  \centering
  \resizebox{\linewidth}{!}{
  \begin{tabular}{llrrrr}
    \toprule
    Setting & & \multicolumn{2}{c}{Unsupervised} & \multicolumn{2}{c}{Linear} \\
    \cmidrule(lr){3-4}
    \cmidrule(lr){5-6}
    Method & Model & Acc. & mIoU & Acc. & mIoU\\
    \midrule
    IIC~\cite{ji2019invariant} & R18+FPN & 21.8 & 6.7 & 44.5 & 8.4 \\
    PiCIE~\cite{cho2021picie} & R18+FPN & 48.1 & 13.8 & 54.2 & 13.9 \\
    PiCIE+H~\cite{cho2021picie} & R18+FPN & 50.0 & 14.4 & 54.8 & 14.8 \\
    \midrule
    DINO~\cite{caron2021emerging} & ViT-S/8 & 28.7 & 11.3 & 68.6 & 33.9 \\
    ACSeg~\cite{li2023acseg} & ViT-S/8 & 16.4 & - & - & - \\
    TransFGU~\cite{yin2022transfgu} & ViT-S/8 & 17.5 & 52.7 & - & - \\
    STEGO + HP~\cite{seong2023hidden} & ViT-S/8 & \textbf{57.2} & 24.6 & \textbf{75.6} & \textbf{42.7} \\
    STEGO~\cite{hamilton2021stego} & ViT-S/8 & 48.3 & 24.5 & 74.4 & 38.3 \\
    \hspace{1.5mm} + \textbf{Ours} & ViT-S/8 & 56.3 & 25.6 & 73.7 & 38.9 \\
    \hspace{1.5mm} + \textbf{Ours w/ 3D-LHP} & ViT-S/8 & 55.1 & \textbf{26.7} & 73.9 & 37.8 \\
    \midrule
    {DINO~\cite{caron2021emerging, hamilton2021stego}} & {ViT-B/8} & {30.5} & {9.6} & {66.8} & {29.4} \\
    DINOSAUR~\cite{seitzer2023dinosaur}* & ViT-B/8 & 44.9 & 24.0 & - & - \\
    STEGO~\cite{hamilton2021stego} & ViT-B/8 & 56.9 & 28.2 & \textbf{76.1} & 41.0 \\
    \hspace{1.5mm} + \textbf{Ours} & ViT-B/8 &  \textbf{58.6}  & \textbf{29.0} & 75.5 & \textbf{41.6} \\
    \bottomrule
  \end{tabular}
  }
  \caption{\textbf{Evaluation on COCO-Stuff 27.} We report results on COCO-Stuff with 27 high-level classes. Overall, our method outperforms STEGO and HP on unsupervised segmentation with the ViT-B/8, while showing competitive performance for the ViT-S/8.
  *Results obtained without post-processing optimization.
  }
  \label{tab:cocostuff27}
  \vspace{-3mm}
\end{table}

\subsection{COCO-Stuff}
We present our evaluation on COCO-Stuff 27 in Table~\ref{tab:cocostuff27}. For the ViT-S/8, our experiments show that \emph{Ours} is able to improve upon STEGO in most metrics, with improved unsupervised accuracy by \textbf{+8.0\%} and unsupervised mIoU increased by \textbf{+1.1\%}. \emph{Ours w/ 3D-LHP} further increases this mIoU delta to \textbf{+1.8\%}, highlighting the effectiveness of our 3D-information propagation strategy in combination with \emph{DepthG}. When comparing our approach to Hidden Positives, a method with more computational overhead, for the ViT-S/8, we show competitive performance for unsupervised accuracy and outperform their approach by \textbf{+1.0\%} on unsupervised mIoU with \emph{Ours} and \textbf{+1.7\%} with \emph{Ours w/ 3D-LHP}.
When using the DINO ViT-B/8 encoder, our approach again outperforms STEGO, as well as all other presented methods on unsupervised metrics. Most notably, we are able to increase the unsupervised mIoU by \textbf{+0.8\%}. 

\subsection{Cityscapes}
We further evaluate our approach on the Cityscapes dataset~\cite{cordts2016cityscapes}, consisting of various scenes from 50 different cities. We follow the training setting from STEGO and, contrary to all other datasets, do not sample point-wise but use the full feature map for our learning process along with the full depth map. As can be seen in Table~\ref{tab:vit-b_cityscapes}, our method significantly outperforms STEGO as well as Hidden Positives. For unsupervised mIoU, while Hidden Positives decreased performance compared to STEGO, we observe that our approach to achieves a \textbf{+2.1\%} increase. Similarly, we report state-of-the-art performance in accuracy, building upon Hidden Positives' already impressive improvements over STEGO and outperforming it by \textbf{+2.1\%}.

\subsection{Potsdam}
Our model is further evaluated on the Potsdam-3 dataset, containing aerial images of the German city of Potsdam. Contrary to the other benchmarks, which contain images in a first-person perspective, Potsdam-3 contains only birds-eye-view images, a perspective that is considered out-of-distribution for the monocular depth estimator. 
Despite this inherent limitation of our approach for aerial data, we are able to demonstrate relatively commendable performance in Table~\ref{tab:vit-s_potsdam} by improving STEGO's performance and reporting state-of-the-art performance for the ViT-S backbone. In contrast, Hidden Positives~\cite{seong2023hidden} use a ViT-B/8, and with roughly twice the parameters as ours, reach an accuracy of 82.4\%. We present a visual overview of the predicted Potsdam depth maps in the appendix.

\begin{table}[!bht]
  \centering
  \begin{tabular}{llrr}
    \toprule
    Method & Model & U. Acc & U. mIoU \\
    \midrule
    IIC~\cite{ji2019invariant} & R18+FPN & 47.9 & 6.4 \\
    PiCIE~\cite{cho2021picie} & R18+FPN & 65.6 & 12.3 \\
    \midrule
    DINO~\cite{caron2021emerging} & ViT-B/8 & 43.6 & 11.8 \\
    STEGO + HP~\cite{seong2023hidden} & ViT-B/8 & 79.5 & 18.4 \\
    STEGO~\cite{hamilton2021stego} & ViT-B/8 & 73.2 & 21.0 \\
    \hspace{1.5mm} + \textbf{Ours} & ViT-B/8 & \textbf{81.6} & \textbf{23.1}  \\
    \bottomrule
  \end{tabular}
  \caption{\textbf{Results on Cityscapes.}
  We report unsupervised accuracy and mIoU on Cityscapes. Our method outperforms both STEGO variants by substantial margins. Notably, our method is the first to improve upon unsupervised mIoU.
  }
  \label{tab:vit-b_cityscapes}
  \vspace{-4mm}
\end{table}
\begin{table}[!thb]
  \centering
  \begin{tabular}{llc}
    \toprule
    Method & Model & U. Acc.  \\
    \midrule
    CC~\cite{isola2015learning} & VGG11 & 63.9   \\ 
    DeepCluster~\cite{caron2018deep} & VGG11 & 41.7  \\ 
    IIC~\cite{ji2019invariant} & VGG11 & 65.1 \\
    \midrule
    DINO~\cite{caron2021emerging, koenig2023uncovering} & ViT-S/8 & 71.3 \\
    STEGO~\cite{hamilton2021stego, koenig2023uncovering} & ViT-S/8 & 77.0 \\
    \hspace{1.5mm} + \textbf{Ours} & ViT-S/8 & \textbf{80.4} \\
    \hspace{1.5mm} + \textbf{Ours w/ 3D-LHP} & ViT-S/8 & \textbf{80.4} \\
    \bottomrule
  \end{tabular}
  \caption{\textbf{Results on Potsdam.}
  We report unsupervised accuracy on the Potsdam dataset. Our method is able to improve upon STEGO. We hypothesize that with a zero-shot depth estimator more suitable for aerial images, the results for our method could further improve.
  }
  \label{tab:vit-s_potsdam}
  \vspace{-3mm}
\end{table}
\begin{table}[!bht]
  \vspace{-2mm}
  \centering
  \begin{tabular}{lcc}
    \toprule
    Method & U. mIoU. & U. Acc \\
    \midrule
    STEGO~\cite{hamilton2021stego} & 24.5 & 48.3 \\
    \hspace{1.5mm} + Depth-Feature Correlation (1) & 24.7 & 51.2 \\
    \hspace{1.5mm} + FPS (2) & 24.6 & 49.1 \\
    \hspace{1.5mm} + 3D-LHP (3) & 25.2 & 48.5 \\
    \midrule
    \hspace{1.5mm} + \textbf{Ours} (1, 2) & 25.6 & \textbf{56.3} \\
    \hspace{1.5mm} + \textbf{Ours w/ 3D-LHP} (1, 2, 3) & \textbf{26.7} & 55.1 \\
    \bottomrule
  \end{tabular}
  \caption{\textbf{Effect of our contributions.} We compare our individual contributions and the combination of all contributions.}
  \label{tab:abl-individual}
\end{table}



\subsection{Qualitative Results}
\begin{figure*}[!thb]
  \centering
  \begin{subfigure}{0.485\textwidth}
  \includegraphics[width=\textwidth, page=3]{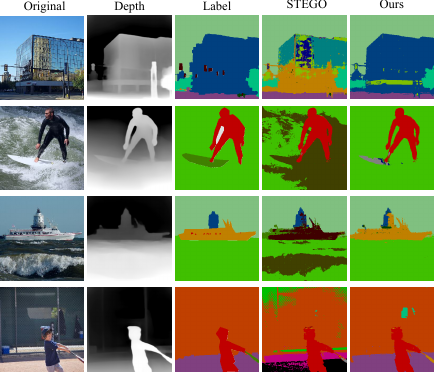}
  \caption{COCO-Stuff}
  \label{subfig:cocostuff}
  \end{subfigure}%
    \hfill
  \begin{subfigure}{0.485\textwidth}
  \includegraphics[width=\textwidth, page=4]{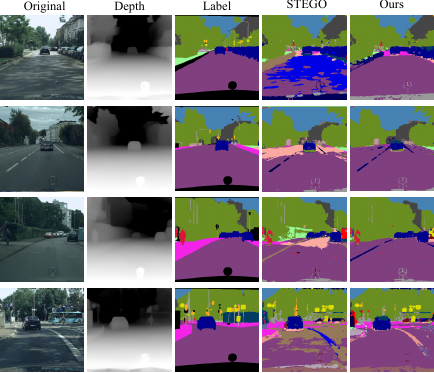}
  \caption{Cityscapes}
  \label{subfig:cityscapes}
  \end{subfigure}
  
  \caption{\textbf{Qualitative results.} We show qualitative differences for plain STEGO compared to STEGO with our depth guidance, using ViT-S models for COCO and ViT-B for Cityscapes. Where STEGO struggles to differentiate instances, our model is able to correct this and successfully separates them for segmentation. In the case of the building in (a), our method alleviates visual irritations from the pixel space and corrects the segmentation of the building. In (b), our model is able to better handle visual inconsistencies from shadows.}
  \label{fig:qualitative}
  \vspace{-3mm}
\end{figure*}

We present qualitative results of our method in Figure~\ref{fig:qualitative} and compare with segmentation maps from STEGO. On multiple occasions, our depth guidance reduces erroneous predictions from the model caused by visual irritations in the pixel space. In the example of the boy with the baseball bat in Figure~\ref{subfig:cocostuff}, false classifications from STEGO are caused by shadows on the ground. Our model is able to correct this. Furthermore, it goes beyond the noisy label and also correctly classifies the glimpse of a plant that can be seen through a hole in the background. This is an indication that our model does not overfit to the depth map, since this visual cue is only observable in the pixel space, but not the depth map.

\section{Ablations}

\paragraph{Individual Influence.}\label{abl:influence}

We investigate the effect of our technical contributions on training our model with a ViT-S/8 backbone on COCO-Stuff 27. Our observations in Table~\ref{tab:abl-individual} show that our \emph{Depth-Feature Correlation} loss itself already improves the performance of STEGO. This improvement is further increased through the use of FPS, which enables us to sample the depth space more meaningfully and therefore encourages more diversity in the depth correlation tensor $\bm{D}_{hw, uv}$. Intuitively, this sampling diversity significantly amplifies our \emph{Depth-Feature Correlation} for aligning the feature space with the depth space. We provide a visual comparison to random sampling in Figure~\ref{fig:sampling} and additional illustrations in the appendix. FPS retrieves more diverse locations and specifically selects locations with depth discontinuities. This naturally benefits the \emph{Depth-Feature Correlation} to learn sharper edges in this area for the output segmentation.
Adding local hidden positives with depth maps further increases the unsupervised mIoU, while slightly lowering the accuracy.

\paragraph{Source Of Depth Maps.}\label{abl:depth-maps}
We investigate the effect of different monocular depth estimators to generate the depth maps used to train our model. In our experiments, we consider three options: The previously mentioned ZoeDepth~\cite{bhat2023zoedepth}, Kick Back \& Relax (KBR)~\cite{spencer2023slowtv} which uses self-supervision to learn depth from Slow-TV videos, as well as MiDaS~\cite{ranftl2020midas}, the base model to ZoeDepth. Empirically, ZoeDepth produces the most accurate zero-shot monodepth results across indoor and outdoor datasets, followed by MiDaS and KBR~\cite{bhat2023zoedepth, spencer2023slowtv}. We provide a qualitative comparison in the appendix.
To evaluate the influence of the depth map quality on the performance of our model, we first generate depth maps for COCO-Stuff 27 using each of the introduced models. We then train our model with a ViT-S/8 backbone and report the results in Table~\ref{tab:abl-depth}. We observe that depth maps from ZoeDepth work best with our method, while the model trained with MiDaS depth has an edge over the KBR counterpart.

\begin{figure}[!thb]
  \centering
   \begin{subfigure}{0.485\linewidth}
  \includegraphics[width=\textwidth, page=7]{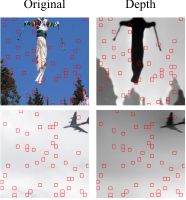}
  \caption{Random Sampling}
  \label{subfig:random-sampling}
  \end{subfigure}%
    \hfill
  \begin{subfigure}{0.49\linewidth}
  \includegraphics[width=\textwidth, page=8]{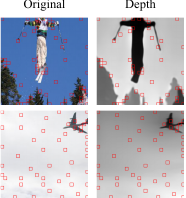}
  \caption{Farthest Point Sampling}
  \label{subfig:fps-sampling}
  \end{subfigure}
   \caption{\textbf{Random vs. Farthest Point Sampling.} We observe that random sampling can miss entire structures like trees in the first top and the plane in the bottom row. In contrast, our method meaningfully samples the depth space and selects locations across the different structures and at depth edges. We show further illustrations of FPS in the appendix.}
   \label{fig:sampling}
   \vspace{-3mm}
\end{figure}

\begin{figure}[!tb]
  \centering
   \includegraphics[width=\linewidth, page=5]{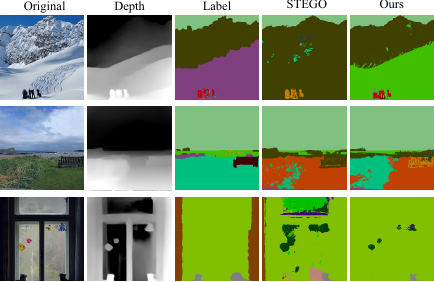}
   \caption{\textbf{Failure cases.} We show cases where our model fails to correctly segment and classify the scene. The top row is a prime example where the difference in depth is correctly distilled, though the model fails to correctly classify the snow region.}
   \label{fig:failure}
   \vspace{-4mm}
\end{figure}

\paragraph{Signal Propagation With Local Hidden Positives.} \label{sec:lhp-abl}
We ablate the implementation of LHP along with different propagation strategies. As described in Section~\ref{sec:lhp}, our method takes advantage of the depth information of the scene to propagate the learning signal to patches which are nearby in 3D space. We further implement utilizing the attention map from the DINO backbone and propagate proportionally to their values i.e., the approach utilized in Hidden Positives~\cite{seong2023hidden}.
While, in Table~\ref{tab:abl-3dlhp}, we show that our approach benefits from signal propagation with 3D-LHP, applying LHP with Attention leads to lower performance. We speculate 3D-LHP is more suitable for our approach since, for a given location, the signal from our \emph{Depth-Feature Correlation} loss is calculated w.r.t. the depth at this sample. Propagating to locations with the same depth does not corrupt this signal, though this can happen with LHP (Attention), since it does not consider depth for propagation.
\vspace{-4mm}
\paragraph{Computational Cost.} \label{sec:cost}
Our method only leads to an insignificant increase in runtime versus the baseline STEGO model, since we solely guide the loss as well as the feature sampling and only for \emph{Ours w/ 3D-LHP} add an additional small segmentation head. In contrast, the competing method Hidden Positives~\cite{seong2023hidden} relies on a computationally more expensive process to select features and introduces an additional segmentation head to fill their task-specific feature pool. To keep our computational overhead low, we make use of a pre-trained monocular depth estimation network with impressive zero-shot capabilities. We consider task specific training of the depth estimator not a necessity, since the model has zero-shot capabilities that generalize well to different scenes and domains. Therefore, in our experiments on a diverse array of scenes, we do not re-train the depth estimator, and consider the additional computational cost for generating the depth maps to be negligible.

\begin{table}[!h]
  \centering
  \begin{tabular}{llcc}
    \toprule
    Method & Trained with & U. Acc. & U. mIoU \\
    \midrule
    ZoeDepth\cite{bhat2023zoedepth} & Sensor Depth & \textbf{56.3} & \textbf{25.6}  \\
    MiDaS\cite{ranftl2020midas} & Sensor Depth & 53.0 & 25.0  \\
    KBR\cite{spencer2023slowtv} & Self-Supervision & 50.6 & 23.1 \\
    \bottomrule
  \end{tabular}
  \caption{\textbf{Different depth map sources.} We experiment with different monocular depth estimators which were trained with either sensor depth or self-supervision. Overall, the model trained with depth maps from ZoeDepth performs best on COCO-Stuff 27.
  }
  \label{tab:abl-depth}
\end{table}

\begin{table}[!h]
  \centering
  \begin{tabular}{clcc}
    \toprule
    LHP & Propagation Strategy & U. Acc. & U. mIoU \\
    \midrule
    \xmark & - & \textbf{56.3} & \underline{25.6} \\
    \cmark & 3D-LHP (Depth) & \underline{55.1} & \textbf{26.7} \\
    \cmark & LHP (Attention) & 52.6 & 24.5 \\
    \bottomrule
  \end{tabular}
  \caption{\textbf{LHP Ablation.} We compare the use of depth and attention for local hidden positives. We find that using depth improves unsupervised mIoU, while we find that using attention does not improve our method.
  }
  \label{tab:abl-3dlhp}
  \vspace{-2mm}
\end{table}

\section{Limitations}
While we have demonstrated our method's effectiveness for many real-world cases, our method's applicability is limited in settings unsuitable for depth estimation, such as slices of CT scans and other medical data domains. Furthermore, the experiments on Potsdam-3 have shown, our method can improve unsupervised semantic segmentation despite suboptimal viewing perspectives for the monocular depth estimator. We assume the scenario of aerial images represents a rare case where our method would profit from a  domain-specific monocular depth estimator.
We also present failure cases of our model in Figure \ref{fig:failure}.

\section{Conclusion \& Future Work}
In this work, we have presented a novel method to induce spatial knowledge of the scene into our model for unsupervised semantic segmentation. We have proposed to correlate the feature space with the depth space and use the 3D information to more meaningfully sample features in a spatially informed way. Furthermore, we have demonstrated that these contributions produce state-of-the-art performance on many real-world datasets and thus foster the progress in unsupervised segmentation.
The applicability of our approach for other tasks is further to be explored, since we hypothesize it can be useful beyond unsupervised segmentation as part of other contrastive processes. We consider this to be a promising direction for future work. Furthermore, it remains to be investigated what kind of information could be useful in domains where depth is not an obviously meaningful signal, like medical data.


{\small
\bibliographystyle{ieee_fullname}
\bibliography{egbib}
}

\newpage

\clearpage
\setcounter{page}{1}

\appendix

\section{Training Details}
\subsection{General Hyperparameters}
We provide the hyperparameters used to train our models. While all models share some common parameters, there are many that can vary for each dataset.  
The hyperparameters in Table \ref{tab:general-hyperparameters} are identical for all models:

\begin{table}[!h]
  \centering
  \begin{tabular}{ll}
    \toprule
    Component \hspace{10mm} & Value \\
    \midrule
    Training Size & $224 \times 224$ \\
    Test Size & $320 \times 320$ \\
    Learning Rate & $5\text{e}^{-4}$ \\
    \bottomrule
  \end{tabular}
  \caption{\textbf{General hyperparameters.}}
  \label{tab:general-hyperparameters}
  \vspace{-3mm}
\end{table}

\subsection{Dataset-specific Hyperparameters}
Table \ref{tab:dataset-hyperparameters} shows the collection of hyperparamers that are specific for our results on various datasets and for various model sizes. We note that Potsdam-3 already undergoes a preprocessing where the images are cropped into $200 \times 200$ sized crops, following \cite{ji2019invariant}. Further, as detailed in the main text, for Cityscapes the entire feature map is used for learning.
\begin{table}[!h]
  \centering
  \resizebox{\linewidth}{!}{
  \begin{tabular}{lcccc}
    \toprule
    \hspace{1cm} Dataset & \multicolumn{2}{c}{COCO-Stuff 27} & Cityscapes & Potsdam \\
    \cmidrule(lr){2-3}
    \cmidrule(lr){4-4}
    \cmidrule(lr){5-5}
    \hspace{1cm} Model & ViT-S & ViT-B & ViT-B & ViT-S\\
    $\downarrow$ Component  & & & & \\
    \toprule
    $\lambda_\text{DepthG}$ & 0.19 & 0.16 & 0.09 & 0.13\\
    $\lambda_\text{self}$ & 0.58 & 0.23 & 0.95 & 0.61\\
    $\lambda_\text{knn}$ & 0.36 & 1.05 & 1.02 & 0.34\\
    $\lambda_\text{random}$ & 0.70 & 0.24 & 0.57 & 0.72\\
    \midrule
    $b_\text{DepthG}$ & 0.03 & 0.03 & 0.03 & 0.14\\
    $b_\text{self}$ & 0.07 & 0.12 & 0.39 & 0.2\\
    $b_\text{knn}$ & 0.02 & 0.21 & 0.25 & 0.09\\
    $b_\text{random}$ & 0.76 & 0.97 & 0.26 & 0.63\\
    \midrule
    Training Steps & 7000 & 7000 & 7000 & 7000\\
    Pointwise Sampling & \cmark & \cmark & \xmark & \cmark \\
    $N$ & 9 & 12 & All & 11\\
    Decay Step & 250 & 300 & 400 & None\\
    Decay Factor & 0.6 & 0.64 & 0.8 & None\\
    Cropping & Five-Crop & Five-Crop & Five-Crop & None \\
    \bottomrule
  \end{tabular}
  }
  \caption{\textbf{Dataset-specific hyperparameters.}}
  \label{tab:dataset-hyperparameters}
  \vspace{-3mm}
\end{table}


\section{Further Ablations}

\subsection{Guidance Variations} \label{sec:depth-effects}
 To explore the functionality of our guidance mechanism, we present further ablations in Table \ref{tab:depth-variations} where we also explore the use of feature maps from the penultimate layer of the monocular depth estimator (MDE) and using image and perspective planes. We further try plugging the ground-truth segmentation maps into the guidance mechanism. We show unsupervised metrics and use the ViT-S/8 config. Our experiments show guidance with depth maps exceeds the placebo effect of using image or perspective planes, and is most effective when utilizing depth maps.

\begin{table}[!h]
    \centering
    \begin{tabular}{lccc}
      \toprule
      & \multicolumn{2}{c}{COCO-Stuff} & Potsdam \\
      \cmidrule(lr){2-3}
      \cmidrule(lr){4-4}
      Guidance & Accuracy & mIoU & Accuracy \\
      \midrule
      STEGO & 48.3 & 24.5 & 77.0  \\
      \midrule
      Depth Map & \textbf{56.3} & \textbf{25.6} & \textbf{80.4}  \\
      MDE Features & 55.9 & 25.4 & 69.3 \\
      Image Plane & 53.3 & 22.8 & 71.1 \\
      Perspective Plane & 52.1 & 23.7 & 67.4 \\
      \midrule
      SemSeg Map & 52.8 & 23.2 & \underline{80.4} \\
      \bottomrule
    \end{tabular}
  \caption{\textbf{Guidance Variations.} We experiment with different ways of guiding our model. In addition to the depth map, we show results for use MDE features, an image and perspective plane, as well as the semantic maps. Our experiments show that depth maps are the most effective guidance modality.}
  \label{tab:depth-variations}
  \vspace{-2mm}
\end{table}

\subsection{Number Of Feature Samples}

\begin{table}[!h]
  \centering
  \resizebox{\linewidth}{!}{
  \begin{tabular}{lcccccccc}
    \toprule
    $N$ & 6 & 7 & 8 & 9 & 10 & 11 & 12  \\
    \midrule
    U. Accuracy & 52.6 & 52.6 & 54.2 & \textbf{56.3} & 53.4 & 54.3 & 54.1 \\
    U. mIoU & 22.2 & 23.0 & 23.8 & \textbf{25.6} & 24.2 & 24.2 & 24.0 \\
    \bottomrule
  \end{tabular}
  }
  \caption{\textbf{Different number of sampled features $N^2$ .} }
  \label{tab:abl-samples}
\end{table}

We ablate varying the number of sampled features $N^2$ for the ViT-S backbone on COCO-Stuff 27. Table \ref{tab:abl-samples} shows the results. For $N=9$, our method obtains the best result. Generally, more samples work better than fewer samples. For $N<8$, our method shows a significant drop in performance. We further find that for the ViT-S model for COCO-Stuff 27, reducing the number of samples during training can lead to a slight gain in performance. There, $N$ is reduced by 1 at every 3000 steps.

\subsection{Guidance Scheduling}

We evaluate the effect of scheduling the impact of our \emph{Depth-Feature Correlation} loss. As detailed in the main text, with our method, we enable to model to get a head start and learn about the rough structure in the scene, to then shift the focus on learning representation from the images as training progresses. Our experiments in Table \ref{tab:abl-scheduling} confirm this. When disabling guidance scheduling, our model's performance deteriorates. 

\begin{table}[!h]
  \centering
  \begin{tabular}{lcc}
    \toprule
    Guidance Scheduling & \xmark & \cmark \\
    \midrule
    Unsupervised Accuracy & 49.4 & \textbf{56.3} \\
    Unsupervised mIoU & 18.8 & \textbf{25.6} \\
    \bottomrule
  \end{tabular}
  \caption{\textbf{STEGO + Ours with and without guidance scheduling.} }
  \label{tab:abl-scheduling}
\end{table}

\subsection{NYUv2 With Ground Truth Depth}
\begin{table}[!h]
    \centering
    \begin{tabular}{lc}
      \toprule
      Depth Source & mIoU \\
      \midrule
      ZoeDepth & 26.1  \\
      Sensor GT & 26.2  \\
      \bottomrule
    \end{tabular}
    \caption{\textbf{Results On NYUv2.} We compare using predicted depth to using ground-truth depth}
    \label{tab:nyuv2}
\end{table}
To compare how our method performs with ground-truth depth vs. predicted depth from ZoeDepth~\cite{bhat2023zoedepth}, we evalute our model on the NYUv2 \cite{silberman2012indoor} semantic segmentation dataset. We report results in Table \ref{tab:nyuv2} and observe that using predicted depth yields similar results as using ground-truth depth.

\subsection{Farthest Point Sampling Visualizations}
\begin{figure}[!h]
  \centering
   \begin{subfigure}{0.485\linewidth}
  \includegraphics[width=\textwidth, page=12]{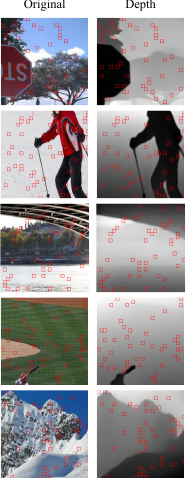}
  \caption{Random Sampling}
  \label{subfig:random-sampling}
  \end{subfigure}%
    \hfill
  \begin{subfigure}{0.484\linewidth}
  \includegraphics[width=\textwidth, page=13]{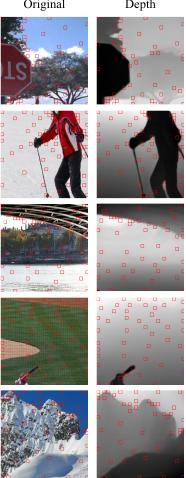}
  \caption{Farthest Point Sampling}
  \label{subfig:fps-sampling}
  \end{subfigure}
   \caption{\textbf{Further examples of Random vs. Farthest Point Sampling.}}
   \label{fig:sampling}
   \vspace{-3mm}
\end{figure}

\begin{figure}[thb]
    \centering
    \begin{subfigure}{\linewidth}
    \includegraphics[width=\textwidth]{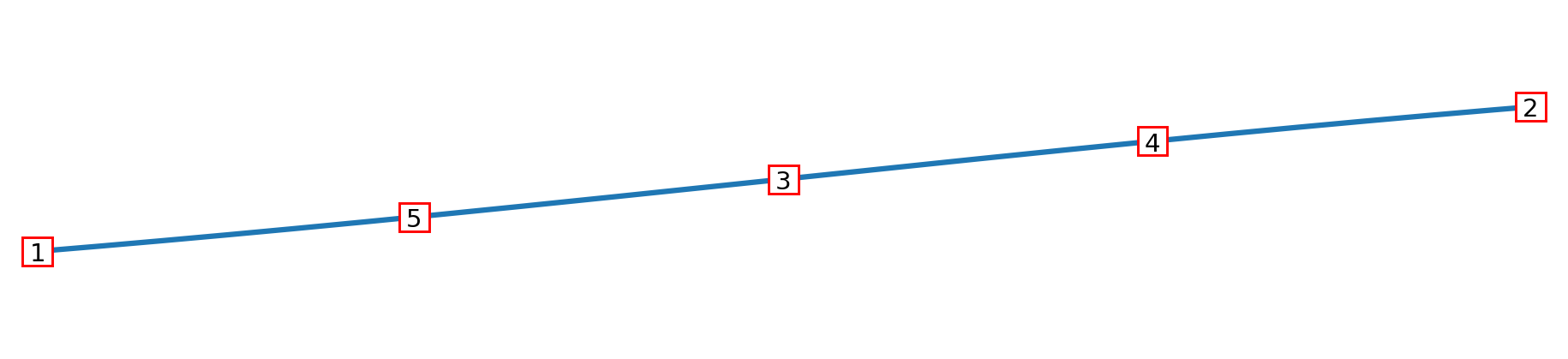}
    \caption{\textbf{Continuous Signal}: Evenly spaced samples}
    \end{subfigure}%
    \\
    \begin{subfigure}{\linewidth}
    \includegraphics[width=\textwidth]{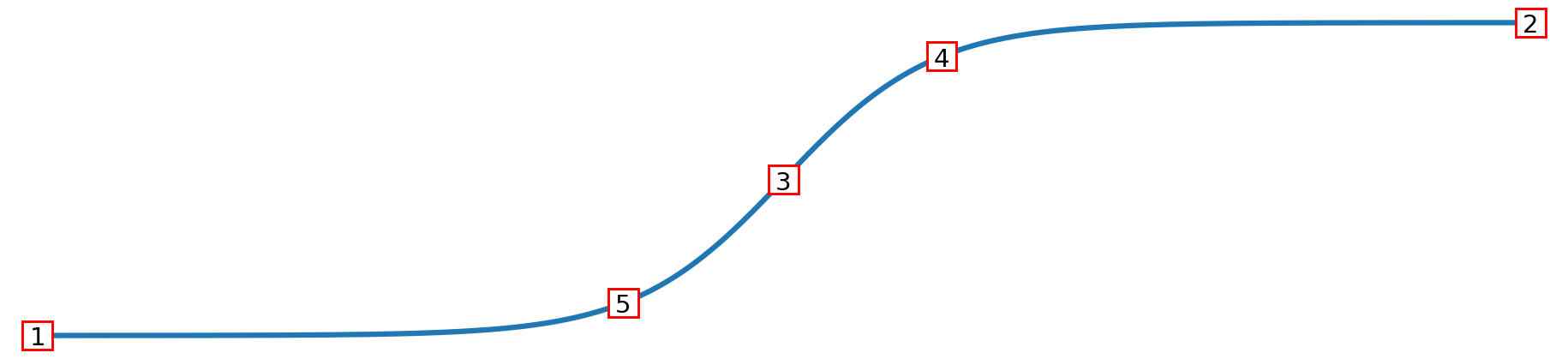}
    \caption{\textbf{Smooth Signal}: Samples concentrate slightly at the gradient}
    \end{subfigure}%
    \\
    \begin{subfigure}{\linewidth}
    \includegraphics[width=\textwidth]{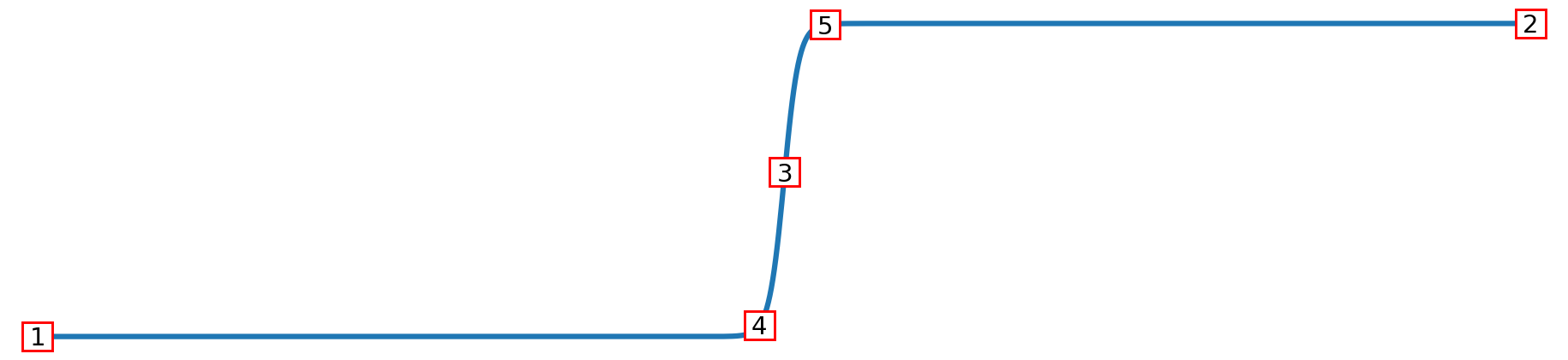}
    \caption{\textbf{Sharp Signal}: Samples concentrate at depth gradient}
    \end{subfigure}
    \caption{\textbf{Visualization of FPS.} We show that samples in FPS converge towards a depth gradient in the signal. The stronger the gradient the more samples are drawn at this region, resulting in meaningful samples for our Depth-Feature Correlation Loss. Numbers denote the order in which the points are sampled.}
    \label{fig:fps-vis-synthetic}
\end{figure}

\begin{figure*}[tb]
    \centering
    \includegraphics[width=\textwidth]{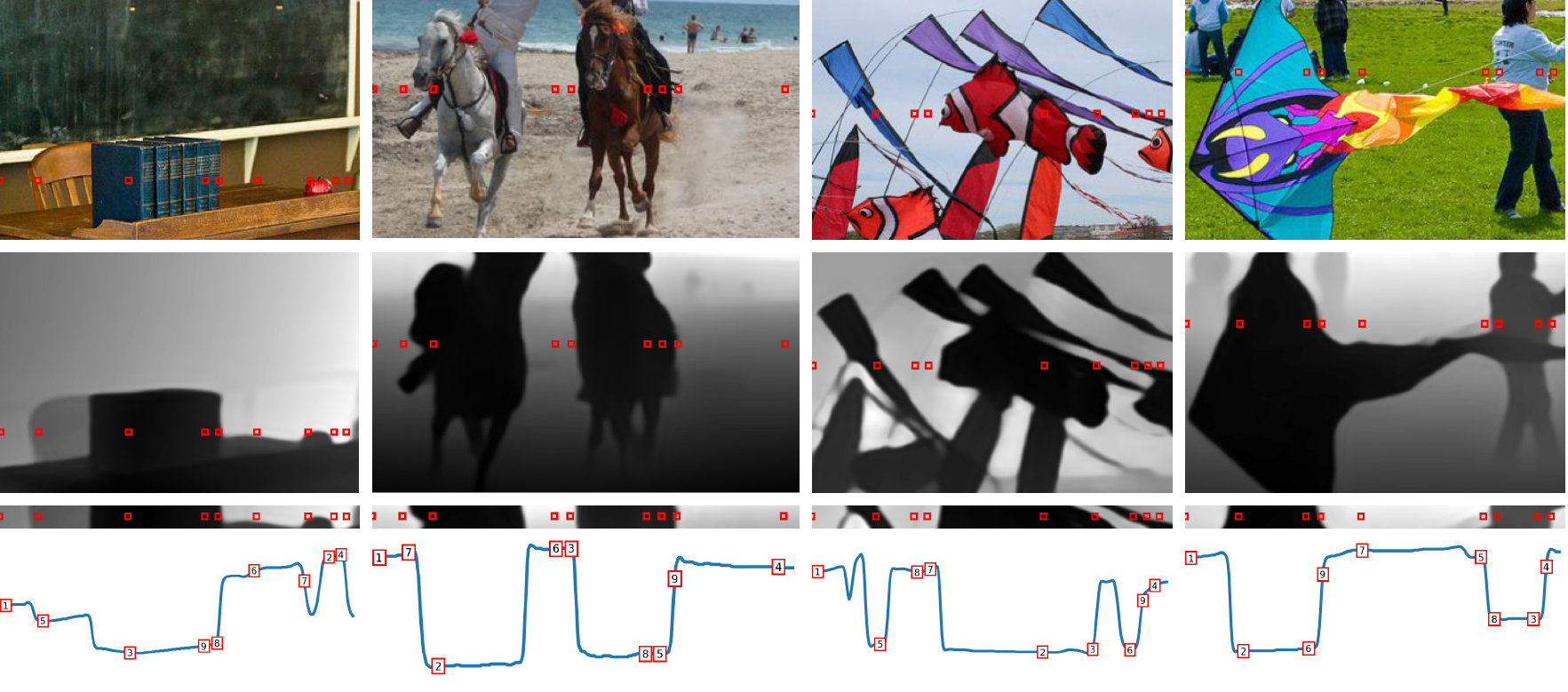}
    \caption{\textbf{Visualization of FPS.} We show how FPS samples the depth space on a selection of images. To show the sapling process, we apply FPS along a line in the image to show how it behaves for sampling the depth space. We project the sampled locations along with the depth gradients onto a 1D plot in the bottom row. It can be observed that FPS samples along the edges of objects and encourages depth diversity in the chosen samples.}
    \label{fig:fps-vis-real}
\end{figure*}

\begin{figure*}[tb]
    \centering
    \includegraphics[width=\textwidth, page=10]{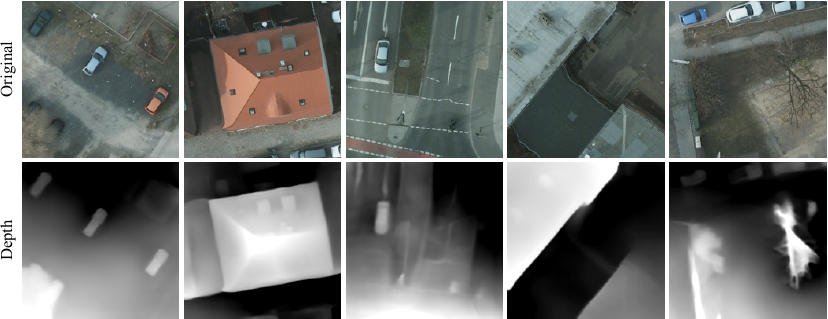}
    \caption{\textbf{Predicted Depth for Potsdam dataset.} We use ZoeDepth \cite{bhat2023zoedepth} to predict on the Potsdam aerial images. As can be observed from the visualization, the predictions can overemphasize the foreground and display a bias of predicting depth gradients towards the bottom of the depth map, despite no significant change in depth.}
    \label{fig:potsdam-zoe}
\end{figure*}

To further underline the importance of FPS for our method, and to provide an intuition for how it samples feature, we show additional sampling visualizations.
In Figure~\ref{fig:fps-vis-real}, we display FPS along a sampled axis in the image and depth map. The depth gradient is displayed below in 1D, along with the samples visualized. Figure~\ref{fig:fps-vis-synthetic} provides an intuition how FPS selects sampling locations along different gradients. For a continuous signal, FPS samples to space evenly, but the sharper the surface becomes, the more samples are concentrated around the gradient. These 3D sampling locations are consequently converted to 2D samples and after conversion, they appear around the edges of objects.
We also show further random sampling vs. FPS examples in Figure~\ref{fig:sampling}.

\section{Qualitative Results}
\subsection{More Results}
We show additional qualitative results in Figure~\ref{fig:more-qualitative}. All results were generated by ViT-S models, also for competitive methods. Throughout all examples, our depth guidance is effective at enabling our model to segment the scene nicely with more consistent surfaces.

\subsection{Comparison to HP}

\begin{figure*}[tb]
    \centering
    \includegraphics[width=\textwidth, page=9]{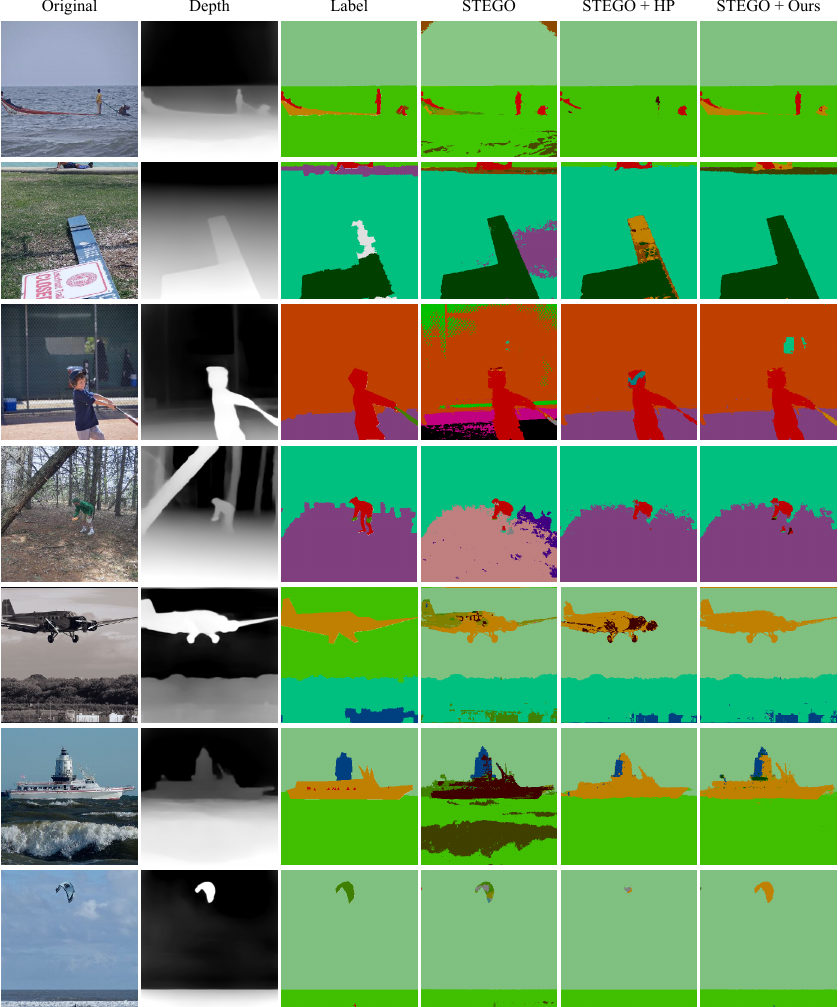}
    \caption{\textbf{More Qualitative Results on COCO-Stuff 27.} We show further qualitative results, adding also a comparison to Hidden Positives.}
    \label{fig:more-qualitative}
\end{figure*}

As part of Figure~\ref{fig:more-qualitative}, we also add qualitative comparisons to Hidden Positives. While their approach significantly improves the performance of STEGO on the shown examples, our method often produces more consistent segmentations for surfaces. For example, in the top row, Hidden Positives fails to segment the boat at all, while our method produces the correct segmentation.

\subsection{Potsdam Depth Predictions}

As mentioned in the main text, we show examples of depth precitions from ZoeDepth \cite{bhat2023zoedepth} in Figure~\ref{fig:potsdam-zoe}. While the predictions have sharp boarders around houses, there are a few cases displayed where the model struggles. For example, in the most left column, it produces a depth gradient towards the bottom of the images, despite the entire parking lot having the same depth. In the center column, the part of the road in the top right-hand corner is predicted as further away, while the red cyclist path appears much closer. Further, in the most right column, the model is irritated by the trees.

\subsection{Source Of Depth Maps}

\begin{figure*}[tb]
    \centering
    \includegraphics[width=\textwidth, page=11]{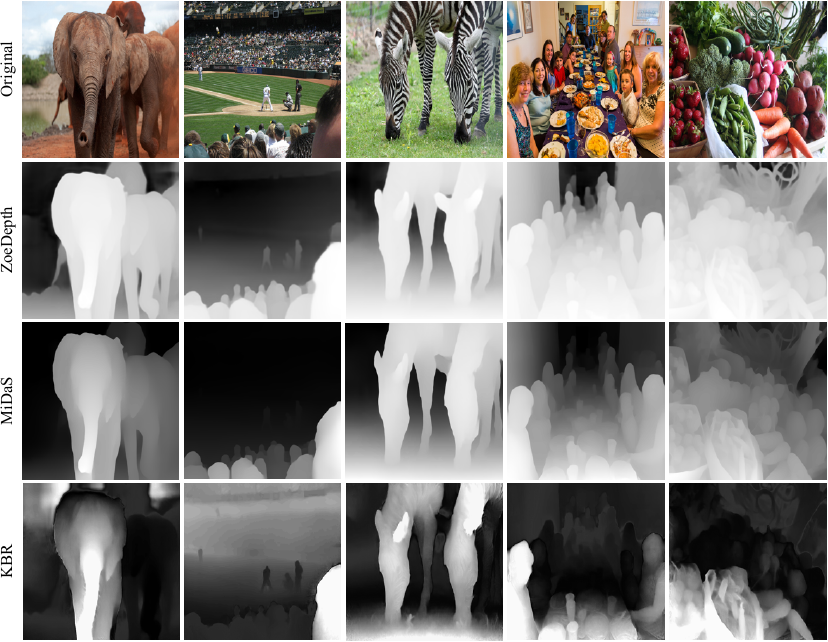}
    \caption{\textbf{Comparison of different monocular depth estimators on COCO-Stuff 27.} Our visualizations qualitatively compares the depth maps predicted by ZoeDepth \cite{bhat2023zoedepth}, MiDaS \cite{ranftl2020midas}, and Kick Back \& Relax \cite{spencer2023slowtv}.}
    \label{fig:depth-comparison}
\end{figure*}

Figure~\ref{fig:depth-comparison} provides a qualitative comparison of the depth maps produced by ZoeDepth \cite{bhat2023zoedepth}, MiDaS \cite{ranftl2020midas} and Kick Back \& Relax \cite{spencer2023slowtv}. All maps are Min-Max normalized. Out of all models, ZoeDepth produces the most consistent depth surfaces, even for complex COCO-Stuff scenes such as crowds. The depth maps from MiDaS are similar but lack detail. Kick Back \& Relax shows impressive results for a self-supervised method, but fails to capture details such as the right zebra's ears in the center column.

\end{document}